%% file: main.tex
\documentclass[runningheads]{llncs}
\usepackage[T1]{fontenc}
\usepackage{graphicx}
\usepackage{caption}
\usepackage{subcaption}
\usepackage{hyperref}
\begin{document}
\title{LLM-Assisted Modeling of Semantic Web-Enabled Multi-Agents Systems with AJAN}
%
%\titlerunning{Abbreviated paper title}
% If the paper title is too long for the running head, you can set
% an abbreviated paper title here
%

\titlerunning{LLM-Assisted Agent Modeling with AJAN}
%\titlerunning{LLM-Assisted Modeling of Semantic Web-Enabled Multi-Agents Systems with AJAN}
% If the paper title is too long for the running head, you can set
% an abbreviated paper title here
%

\author{Hacane Hechehouche\inst{1} \and Andr\'e Antakli\inst{2} \and Matthias Klusch\inst{2}}
\authorrunning{H. Hechehouche et al.}
% First names are abbreviated in the running head.
% If there are more than two authors, 'et al.' is used.
%
\institute{Daimler Buses GmbH\\
          Carl-Zeiss-Straße 2, Neu-Ulm, Germany\\
          hacane.hechehouche@daimlertruck.com\\
          \and 
          German Research Center for Artificial Intelligence (DFKI) \\
           Saarland Informatics Campus, Saarbruecken, Germany\\
          \{andre.antakli, matthias.klusch\}@dfki.de\\
          }
\maketitle              % typeset the header of the contribution
\begin{abstract}
There are many established semantic Web standards for implementing multi-agent driven applications. 
The AJAN framework allows to engineer multi-agent systems based on these standards. In particular, agent knowledge is represented in 
RDF/RDFS and OWL, while agent behavior models are defined with Behavior Trees and SPARQL to access and manipulate this knowledge. 
However, the appropriate definition of RDF/RDFS and SPARQL-based agent behaviors still remains a major hurdle not only for 
agent modelers in practice. For example, dealing with URIs is very error-prone regarding typos and dealing with complex SPARQL 
queries in large-scale environments requires a high learning curve. 
In this paper, we present an integrated development environment to overcome such hurdles of modeling AJAN agents and 
at the same time to extend the user community for AJAN by the possibility to leverage Large Language Models for 
agent engineering. 

%\keywords{First keyword  \and Second keyword \and Another keyword.}
\end{abstract}

\input{sections/introduction}

\input{sections/ide_short}

\input{sections/llm_short}

\input{sections/applications}

\input{sections/relatedwork_short}

\input{sections/conclusion}

% == ACKNOWLEDGMENTS ======================================
\begin{credits}
\subsubsection{\ackname} 
This work has been funded by the German Ministry for Research, Technology 
and Space (BMFTR) in the project MOMENTUM (01IW22001), 
and the European Commision in the project InnovAIte (09I02-03-V01-00029).
\end{credits}

%
% ---- Bibliography ----
%
% BibTeX users should specify bibliography style 'splncs04'.
% References will then be sorted and formatted in the correct style.
%
% \bibliographystyle{splncs04}
% \bibliography{mybibliography}
%

\end{document}

%% file: sections/introduction.tex
\section{Introduction}

\noindent
The AJAN (Accessible Java Agent Nucleus) multi-agent engineering framework \cite{antakli2023ajan} has already been applied across a wide range of domains, demonstrating the versatility of its approach to modeling agents entirely in RDF. A central element of AJAN is its use of {\em SPARQL Behavior Trees (SBTs)} to describe agent behavior. This approach leverages SPARQL—a powerful query language for RDF data—to access and manipulate agent knowledge, while also relying on the modular structure of Behavior Trees, originally developed in the gaming industry and now widely adopted in robotics for high-level behavior programming. The use of SBTs enables not only reusability of reactive behavior modules but also seamless integration of additional AI techniques.
However, defining agents and their behaviors entirely in RDF and SPARQL poses a significant challenge—even for experts in the Semantic Web field—due to the technical complexity and lack of intuitive tooling. Recent developments in generative AI, particularly with Large Language 
Models (LLM), enable the use of the most natural form of communication—natural language—for the creation of MAS-driven applications. When considering AJAN behaviors described in RDF while combining Behavior Trees (LLM approaches: \cite{cao2023robot,jin2022integrating}) and SPARQL (LLM approaches: \cite{rony2022sgpt,kovriguina2023sparqlgen}), a holistic solution must be found that is modular and applicable to dynamic domains.

\noindent
To address this challenge, we developed a {\em web-based graphical editor for AJAN} that combines graphical and tabular interfaces with core features of modern integrated development environments (IDEs). In particular, recent advances in the use of LLMs have been incorporated into the editor to offer a natural, user-friendly interface for modeling agent behavior and interacting directly with an agent acting in a dynamic environment. That enables even non-experts to engage with the AJAN framework for agent modeling more effectively and lowers the entry barrier for agent modeling. This paper is structured as follows: Section 2 introduces the AJAN-Editor, followed by its LLM-based NLP extension in Section 3. Section 4 presents an application example. Related work is discussed in Section 5, and conclusions are drawn in Section 6.
\vspace{-0.2cm}

%% file: sections/ide_short.tex
\section{Hypermedia Agent Modeling}

\begin{figure}[ht]
\vspace{-0.8cm}
    \centering
    \includegraphics[width=0.8\linewidth]{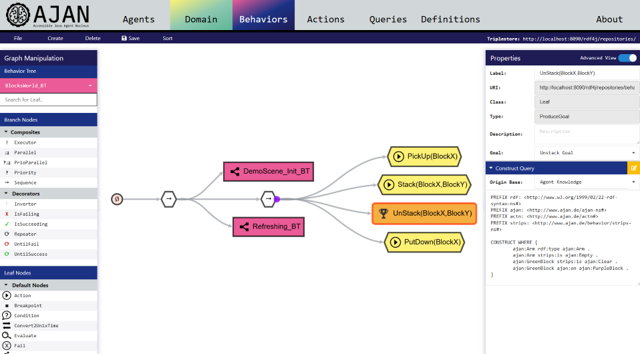}
    \vspace{-0.2cm}
    \caption{Behavior editor to model SPARQL-BTs within the AJAN-Editor}
    \label{fig:editor-bt}
    \vspace{-0.7cm}
\end{figure}

\noindent 
The AJAN-Editor, open-source available on GitHub\footnote{AJAN-Editor: https://anonymous.4open.science/r/AJAN-editor-0E30/}, 
is a web-based graphical interface for modeling, executing, and debugging agents in the AJAN framework. 
It addresses the challenge of authoring RDF-based agent models and SPARQL Behavior Trees (SBTs) by providing an interactive environment 
that lowers the entry barrier for users. Developed with \texttt{Ember.js}\footnote{EmberJS: \url{https://emberjs.com/}}, 
\texttt{Node.js}\footnote{NodeJS: \url{https://nodejs.org/en}} and \texttt{Cytoscape.js}\footnote{Cytoscape: \url{https://js.cytoscape.org/}}, the editor connects to the AJAN-Service and RDF triplestores (e.g., \texttt{RDF4J} or \texttt{GraphDB}), 
enabling the creation and management of agent templates including endpoints, events, goals, and SBTs. 
The editor features visual and tabular tools for constructing event-driven agent behavior via SPARQL queries, and supports live monitoring, execution tracing, and model updates; it also facilitates collaborative development and component reuse.

\subsection{The AJAN Framework}

\noindent 
The AJAN service forms the runtime core of the AJAN framework, enabling execution of Semantic Web-enabled agents and multi-agent systems through RESTful Linked Data (LD) interfaces. Agent models and behaviors are stored in RDF and manipulated via SPARQL. Agents follow the BDI paradigm, with RDF-based knowledge, SPARQL \texttt{ASK}-defined goals, and behavior trees expressed as SPARQL-BTs. These SBTs use SPARQL \texttt{ASK}, \texttt{SELECT}, \texttt{UPDATE}, and \texttt{CONSTRUCT} for decision-making and communication. The modular SBT structure allows for reuse and adaptation of agent logic. For more details, we refer the reader to \cite{antakli2023ajan}.

\subsection{AJAN-Editor: Overview}

\begin{figure}[h!]
\vspace{-0.5cm}
    \centering
    \includegraphics[width=0.9\linewidth]{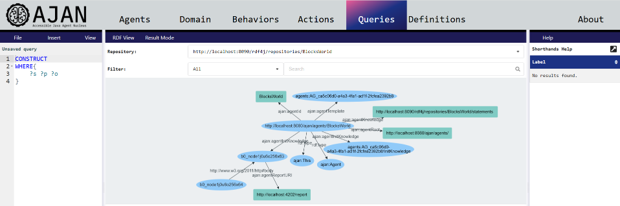}
    \vspace{-0.2cm}
    \caption{Queries tab to model queries and access RDF triple stores}
    \label{fig:editor-KB}
    \vspace{-0.5cm}
\end{figure}

\noindent
The \textit{Agents tab} of the AJAN-Editor enables the user to configure and instantiate agents from RDF-defined agent templates, each specifying behaviors, events, goals, and endpoints. Users can assign URIs as agent endpoints, link SBTs to events and goals, and manage templates, while the respective interface supports validation, import/export, and modular deployment. 
The \textit{Behaviors tab} (cf. Fig. \ref{fig:editor-bt}) offers a graphical editor for modeling SBTs using drag-and-drop of SBT nodes 
like composites, decorators, and leaves. This includes support for zooming, node linking, and SPARQL validation. A dynamic properties panel facilitates configuration of node-specific parameters and SPARQL queries. 
The \textit{Actions tab} supports the definition of AJAN agent actions, including semantic pre- and post-conditions and REST bindings for synchronous or asynchronous Linked Data (LD) interaction. 
The \textit{Queries tab} integrates a SPARQL editor (\texttt{Ace}-based\footnote{Ace Editor: \url{https://ace.c9.io/}}) with resizable panes 
and tabbed tools for management, code insertion, and help. It enables creation, testing, and reuse of SPARQL queries within the agent development process, and visualizes query results in a tabular or graphical form (cf. Fig. \ref{fig:editor-KB}). The queries can be applied to any W3C-standardized SPARQL endpoint\footnote{SPARQL 1.1 Protocol: \url{https://www.w3.org/TR/sparql11-protocol/}}, like the AJAN agent knowledge bases (KBs).

\begin{figure}[h!]
\vspace{-0.5cm}
    \centering
    \includegraphics[width=0.9\linewidth]{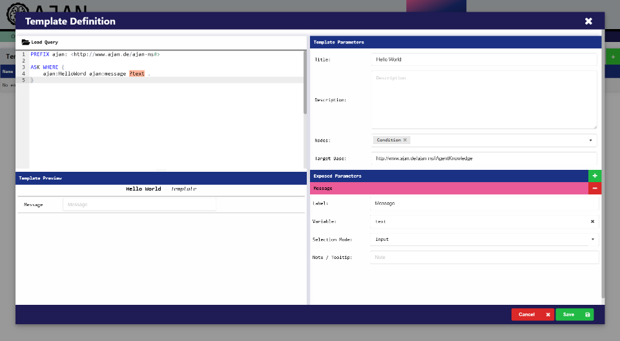}
    \vspace{-0.2cm}
    \caption{Templates view to create parametrized SPARQL templates and UIs}
    \label{fig:editor-templates}
    \vspace{-0.9cm}
\end{figure}

\noindent 
The \textit{Definitions tab} allows the user to manage reusable RDF vocabularies, SPARQL snippets, and templates. 
In the \textit{Templates section} (cf. Fig. \ref{fig:editor-templates}), the user may define parameterized SPARQL queries with custom input forms including text fields, dropdowns, and SPARQL-driven menus. A preview interface shows the resulting UI, enabling verification and reuse across agent models. These templates with their UIs can be used for SBT nodes and hide complex SPARQL queries behind a user-friendly view in the respective node property in the \textit{Behaviors tab}.

\subsection{Monitoring, Testing and Learning}

\begin{figure}[h!]
\vspace{-0.5cm}
    \centering
    \includegraphics[width=0.9\linewidth]{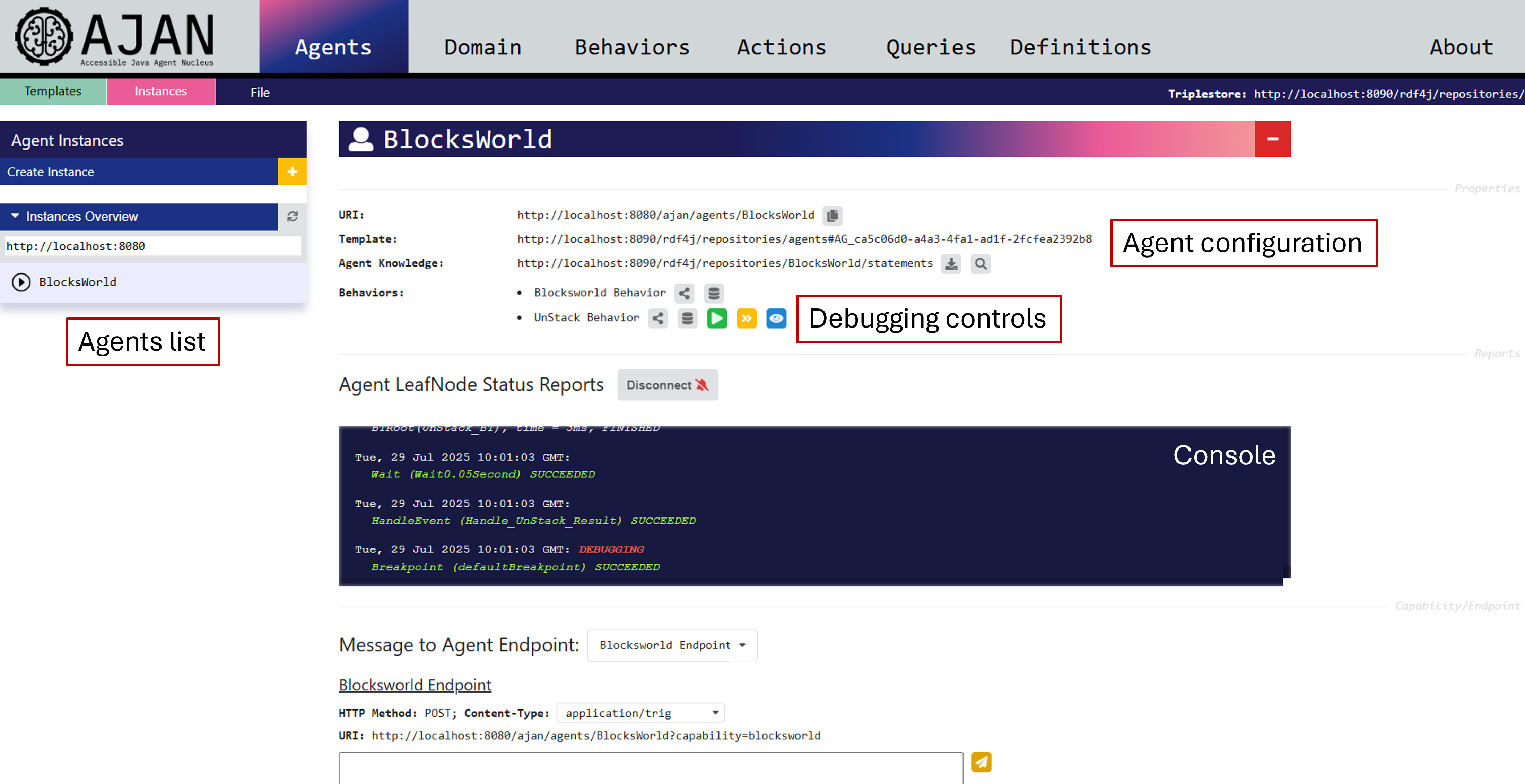}
    \vspace{-0.2cm}
    \caption{Agent instances view to monitor initiated AJAN agents}
    \label{fig:editor-live}
    \vspace{-0.5cm}
\end{figure}

\begin{figure}[h!]
\vspace{-0.5cm}
    \centering
    \includegraphics[width=0.8\linewidth]{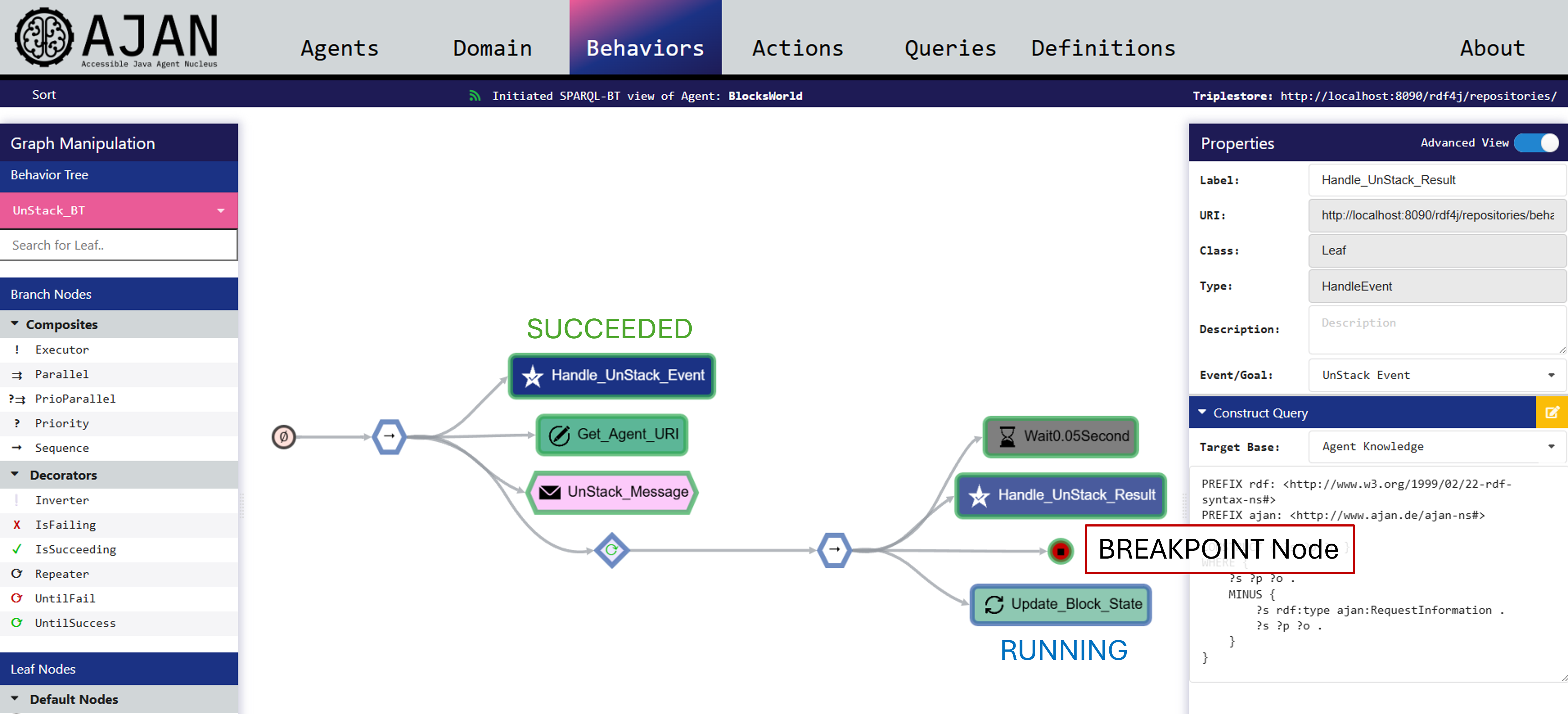}
    \vspace{-0.2cm}
    \caption{Live view of executed SBTs, indicating node states using colored borders}
    \label{fig:editor-live-bt}
    \vspace{-0.5cm}
\end{figure}

\noindent 
The AJAN-Editor provides an interface for initializing and executing agents based on predefined templates (cf. Fig. \ref{fig:editor-live}). Accessible via the \textit{Agents tab}, it offers an overview of all instantiated agents. The instance view displays runtime details, including the assigned template and currently executing behaviors. These behaviors can be monitored in real time through the live view (cf. Fig. \ref{fig:editor-live-bt}), and the agent's KB can be queried and visualized in the \textit{Queries tab}.

\noindent 
A log view shows the execution status of individual SBT nodes. The interface also supports interaction via agent REST endpoints. When a \texttt{BREAKPOINT} node is added in a SBT and reached during execution, the editor enters the debugging mode that allows the
stepwise execution of the SBT, display of the node status, and running of associated SPARQL queries against the agent KB.

\begin{figure}[h!]
\vspace{-0.5cm}
    \centering
    \includegraphics[width=0.8\linewidth]{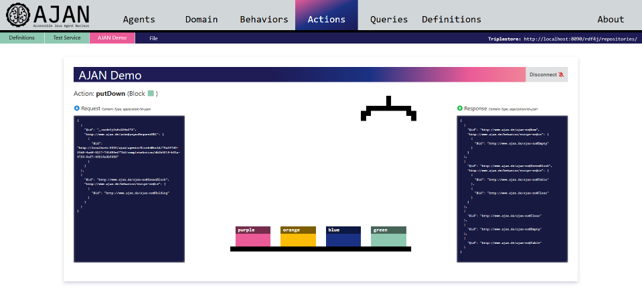}
    \vspace{-0.2cm}
    \caption{World scenario demo environment to learn the modeling of agents}
    \label{fig:editor-demo}
    \vspace{-0.5cm}
\end{figure}

\noindent 
For testing agents and verifying Linked Data-based communication, the \textit{Actions} tab provides tools to monitor inbound agent messages and define respective synchronous or asynchronous LD response messages. 
The same tab includes a demo environment for the user, that shows a basic example of agent modeling with AJAN 
(cf. Fig. \ref{fig:editor-demo}) in a classic Blocks World scenario, where an agent manipulates four colored RDF-described blocks via a LD endpoint. 
Using predefined actions like \texttt{PickUP}, \texttt{PutDown}, \texttt{Stack}, and \texttt{UnStack}, the agent rearranges 
the blocks to achieve specified goals. These actions are visualized via an animated robotic arm, and all message exchanges 
are shown to illustrate behavior execution. Combined with the live view, debugging tools, the AJAN Wiki\footnote{AJAN-Service Wiki: \url{https://github.com/aantakli/AJAN-service/wiki}}, 
and a pre-configured Blocks World agent behavior (cf. Fig. \ref{fig:editor-bt}), this environment offers a complete learning toolkit for new users.

%% file: sections/llm_short.tex
\section{LLM-Assisted Modeling of Agents}

% paper title "with AJAN": HYENA is part of AJAN ;-)

The LLM-based user interface (available on GitHub\footnote{AJAN-Editor NLP-Extension: \url{https://anonymous.4open.science/r/HYENA-69FF}}) for the Semantic Web-enabled AJAN framework is modular and scalable, supporting seamless human-agent interaction. Designed for extensibility, it integrates with AJAN’s RDF-based agent model and enables workflows such as behavior generation, SPARQL querying, and semantic search of the AJAN Wikis. 

\subsection{LLM-Based User Interface: Overview}

\begin{figure}[ht!]
\vspace{-0.5cm}
    \centering
    \includegraphics[width=1\linewidth]{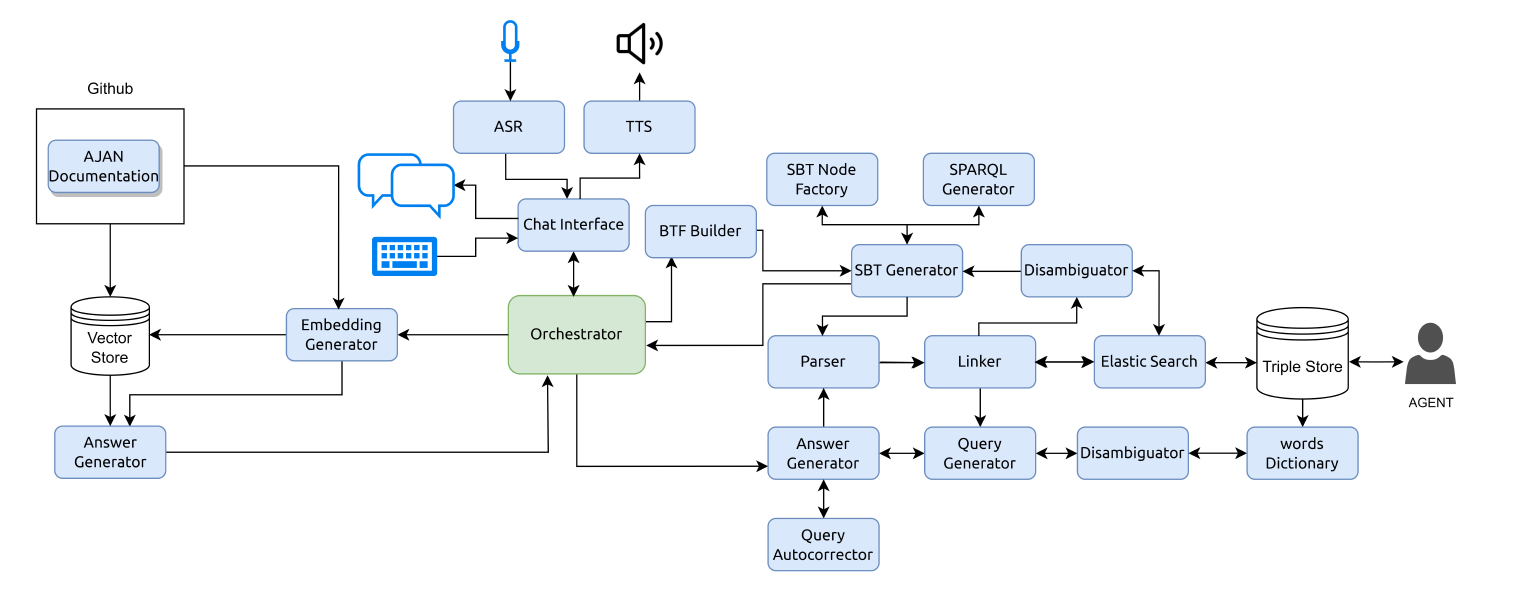}
    \vspace{-0.5cm}
    \caption{Overview of the LLM-based AJAN modeling interface}
    \label{fig:HYENA}
    \vspace{-0.5cm}
\end{figure}

The system architecture (cf. Fig. \ref{fig:HYENA}) comprises two main component classes: common and specialized. 
Common components are shared across workflows and form the foundation of input processing. These components include the \textit{Parser} for syntactic and semantic analysis, the \textit{Linker} for mapping entities to RDF resources, and the \textit{Disambiguator} for resolving ambiguities. In this context, the used \texttt{Elasticsearch} engine\footnote{\url{https://www.elastic.co}} supports fuzzy matching and retrieval from the agent KB. The user-facing layer includes chat and orchestration modules. The \textit{ASR} (Automatic Speech Recognition) transcribes spoken input, while TTS (Text-to-Speech) returns audio responses. The central \textit{Orchestrator} routes input to the appropriate workflow and coordinates response generation. In the SPARQL query workflow, the \textit{Query Generator}, \textit{Autocorrector}, and \textit{Answer Generator} work together to translate intent into executable queries. A \textit{Word Dictionary} supports personalized, consistent vocabulary usage. The SBT workflow allows users to define agent behaviors via natural language, using the \textit{BTF Builder} to construct a high-level tree, the \textit{SBT Generator} to populate it, and the \textit{SBT Node Factory} to instantiate elements. The semantic search workflow includes an \textit{Embedding Generator}, \textit{Vector Store}, and \textit{Answer Generator} for retrieving and synthesizing content from AJAN documentation.

\subsection{Chat Interface and Orchestration}
\label{sec:chat-orchestration}

\noindent
User interaction is structured along two dimensions: \textit{interaction modalities}, which define how users communicate with the system (text or voice), and \textit{interaction modes}, which describe the operational context (offline or online). The Natural Language Processing (NLP) interface for AJAN supports both text and voice input. In text mode, users enter instructions directly into the interface, which are passed to the Orchestrator for semantic processing and workflow selection. For voice input, the NLP interface employs an Automatic Speech Recognition (ASR) pipeline: user speech is recorded via microphone and digitized using the open-source \texttt{PyAudio} library\footnote{\url{https://people.csail.mit.edu/hubert/pyaudio}}. The audio is then transcribed by OpenAI’s \texttt{Whisper} model\footnote{\url{https://github.com/openai/whisper}}, which offers high accuracy across diverse acoustic conditions. The resulting text is forwarded to the NLP engine, enabling seamless multimodal interaction. The NLP extension operates in two modes: \textit{offline} and \textit{online}.

\begin{figure} [ht]
\vspace{-0.5cm}
    \centering
    \includegraphics[width=0.9\linewidth]{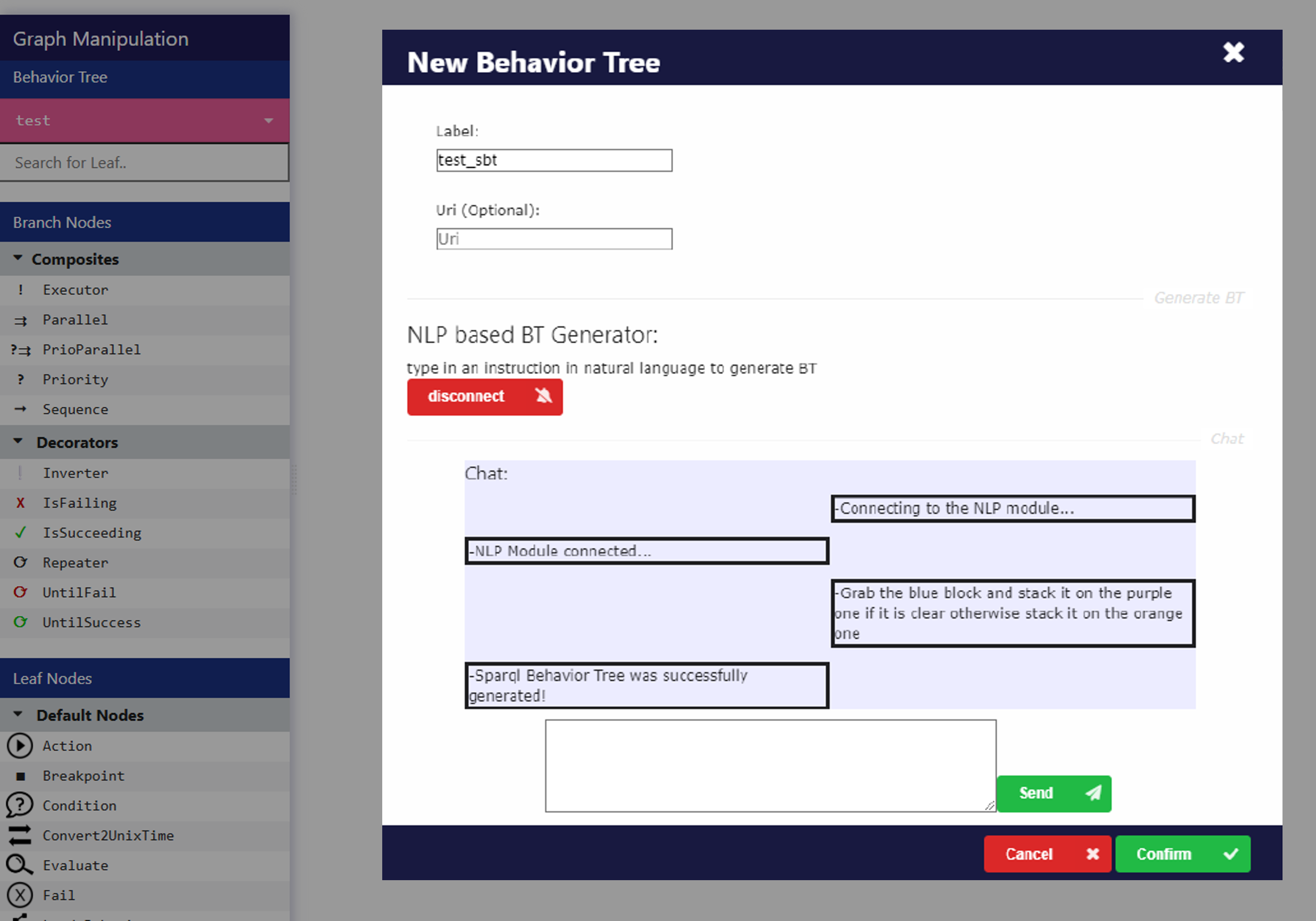}
    \caption{Interface to generate SPARQL-BTs using a LLM-driven chat interface}
    \label{fig:HYENA-offline}
    \vspace{-0.5cm}
\end{figure}

\begin{itemize}
    \item \textbf{Offline Mode:} In offline mode, users interact within the AJAN-Editor (cf. Fig.~\ref{fig:HYENA-offline}, the respectife generated SBT can be seen in Fig. \ref{fig:generated-sbt}) via a chat window in the \textit{Behaviors} tab. Messages are transmitted to the NLP interface through a WebSocket connection. The system responds with a generated SPARQL Behavior Tree, which is visualized in the editor and can be modified, saved, or executed. This mode is suited for agent development and debugging.
    \item \textbf{Online Mode:} Online mode enables direct interaction with agents outside the AJAN-Editor. Users communicate through a console interface, receiving responses in text or synthesized speech via \texttt{PyTTS}\footnote{\url{https://github.com/Kyubyong/pytts}}. Commands result in SBTs sent to and executed by instantiated agents via REST, while queries trigger SPARQL generation and execution against the agent’s RDF knowledge base. This mode emphasizes conversational simplicity while abstracting internal execution details.
\end{itemize}

\subsubsection{Orchestration Logic}

\begin{figure}
\vspace{-0.5cm}
\centering
    \begin{subfigure}[b]{0.6\textwidth}
        \centering
        \includegraphics[width=\textwidth]{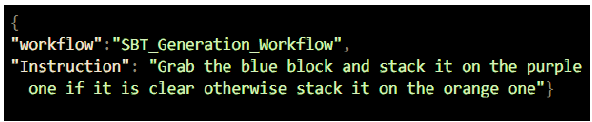}
        \caption{Derived workflow by the Orchestrator}
        \label{fig:orchestration_example}
    \end{subfigure}
    \hfill
    \begin{subfigure}[b]{0.38\textwidth}
        \centering
        \includegraphics[width=\textwidth]{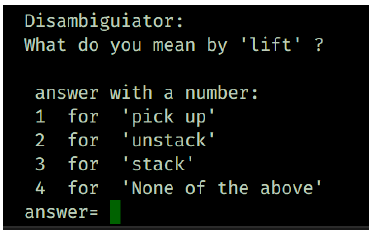}
        \caption{Disambiguation process}
        \label{fig:disambiguation_example}
    \end{subfigure}
    \vspace{-0.2cm}
\caption{Example data of the Orchestrator and Disambiguator}
\label{fig:orch_disam}
\vspace{-0.5cm}
\end{figure}

\noindent
The \textit{Orchestrator} serves as the central controller, interpreting the user input, selecting the appropriate processing workflow, and managing system responses. Upon receiving user input—either transcribed speech or typed text—it performs workflow classification to determine the suitable pipeline. Workflow classification is delegated to the GPT-3.5-turbo-16k-0613 model\footnote{\url{https://platform.openai.com/docs/models/gpt-3-5}}. A prompt defining the context and three possible workflows—SBT generation, SPARQL querying, and semantic search—is combined with the user input. The model returns a structured JSON object indicating the most likely workflow 
(cf. Fig.~\ref{fig:orchestration_example}). Based on this output, the Orchestrator triggers one of the following:
\begin{itemize}
    \item \textit{SPARQL Query Workflow} for simulation-specific queries (see Sect. \ref{sec:sparql-generation})
    \item \textit{SBT Generation Workflow} for behavior-related commands (see Sect. \ref{sec:sbt-factory})
    \item \textit{Semantic Search Workflow} for documentation lookup (see Sect. \ref{sec:semantic-search})
\end{itemize}
It then awaits the workflow result and returns a formatted natural language response according to the current interaction mode. Ambiguous inputs are resolved via the \textit{Disambiguator} module. For instance, if a verb like “lift” is unrecognized in the agent’s ontology, the system prompts for clarification (cf. Fig.~\ref{fig:disambiguation_example}). In case of errors—such as network issues or failed SBT generation—the Orchestrator provides user-friendly feedback. A structured conversation history is maintained to ensure context-aware interaction.

\subsubsection{System Responses}

\noindent
System responses vary by input type. Commands return feedback on SBT generation and execution, while questions are answered using the agent’s knowledge base or documentation. Ambiguous inputs prompt clarifications, and errors result in diagnostic messages. Output is provided as text and optionally via synthesized voice. The \texttt{PyTTS} library supports multiple voice options, enabling adaptation to different user preferences and accessibility needs.

\subsection{LLM-based SPARQL Query Generation}
\label{sec:sparql-generation}

\noindent
Answering user questions about the agent or its environment involves transforming natural language input into executable SPARQL queries evaluated against the agent’s RDF-based knowledge base. This process is both linguistically and semantically complex, requiring several dedicated components and sub-processes.

\subsubsection{Understanding User Queries}
\begin{figure}
\vspace{-0.5cm}
    \centering
    \includegraphics[width=1\linewidth]{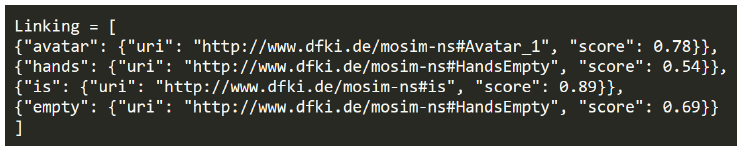}
    \caption{Example output of the linking process}
    \vspace{-0.2cm}
    \label{fig:linking}
    \vspace{-0.9cm}
\end{figure}

\noindent
Accurate query generation begins with interpreting user intent, achieved through three main steps: syntactic parsing, semantic linking, and disambiguation. While \textit{Parsing}, the system first tokenizes the user input and constructs a dependency parse tree using the \texttt{SpaCy} library\footnote{\url{https://www.spacy.io}}. This identifies parts of speech and syntactic roles (e.g., subjects, modifiers), providing a structural representation of the question. The parsed structure helps identify relevant entities and predicates. During \textit{Linking}, identified entities and relations are mapped to RDF resources by comparing them to the ontology used by the agent (contains terminological knowledge and statements about the agent's environment), using specifically \texttt{rdfs:label} values given in the RDF-data. A fuzzy matching algorithm (Levenshtein distance) ranks similarity scores, while Elastic Search provides efficient indexed search. The linking module, adapted from the \texttt{EARL} framework~\cite{dubey2018earl}, returns candidate URIs prioritized by confidence. Figure~\ref{fig:linking} illustrates the output for an example input. If multiple candidates match a user term or confidence is low (threshold set to 0.5), the Disambiguator module prompts the user to select the correct URI. These choices are stored in a synonym dictionary for future reuse, improving personalization and recall.

\subsubsection{Translating Questions into SPARQL Queries}
\begin{figure}[ht!]
\vspace{-0.5cm}
    \centering
    \includegraphics[width=0.9\linewidth]{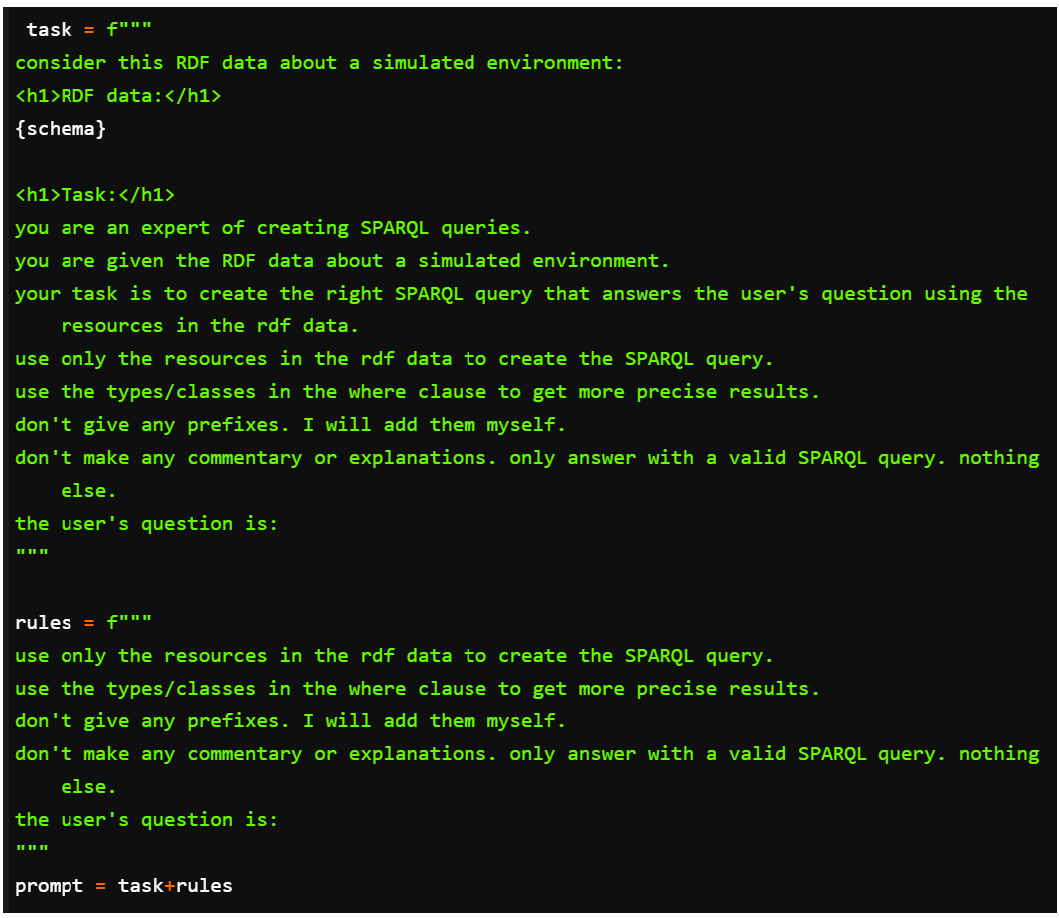}
    \caption{Prompt to generate SPARQL ASK queries using filtered RDF data}
    \vspace{-0.3cm}
    \label{fig:sparql-prompt}
    \vspace{-0.9cm}
\end{figure}

\noindent
After resolving the entities and relations, the system constructs a SPARQL query that corresponds to the user’s question. To minimize errors, Bloom filters are applied to restrict predicate-object combinations to valid patterns observed in the RDF data. This filtering step reduces the likelihood of generating syntactically correct but semantically invalid queries. The filtered ontology, along with a task description and user input, is passed to the GPT-3.5-turbo-16k-0613 model\footnote{\url{https://platform.openai.com/docs/models}} using the prompt depicted in Figure \ref{fig:sparql-prompt}. The model then produces a tailored SPARQL query, either as a \texttt{SELECT} or \texttt{ASK} query, depending on the detected intent. Figure~\ref{fig:condition-node} illustrates in the AJAN-editor a generated query corresponding to the user question \textit{“Is the purple block clear?”}. This generative approach proves effective for well-formed queries but may struggle with lengthy or ambiguous input.

\subsubsection{SPARQL Execution and Error Handling}
\begin{figure}
\vspace{-0.5cm}
    \centering
    \includegraphics[width=1\linewidth]{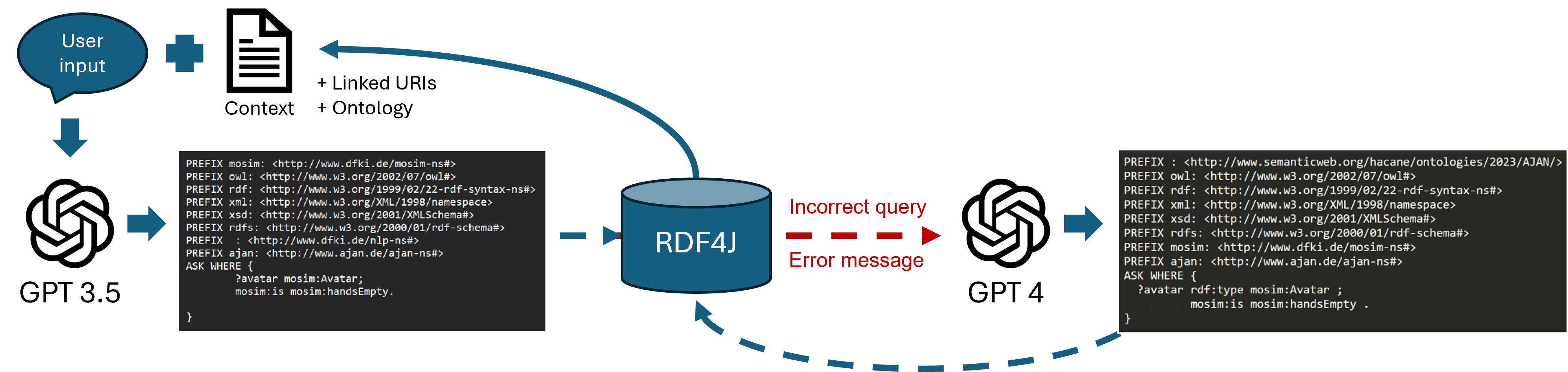}
    \vspace{-0.5cm}
    \caption{SPARQL execution and error handling}
    \label{fig:sparql-autocorrect}
    \vspace{-0.5cm}
\end{figure}

\noindent
The generated query is executed against the agent’s KB via the RDF4J REST API\footnote{RDF4J API: \url{https://rdf4j.org/documentation/reference/rest-api/}}. The results are then parsed and forwarded to the \texttt{Answer Generator}. If execution fails or returns no data, the \texttt{Query Autocorrector} is activated, as shown in Figure \ref{fig:sparql-autocorrect}. This module sends the faulty query, its error message, and surrounding context to the GPT-4-32k-0314 model\footnote{\url{https://platform.openai.com/docs/models}}. GPT-4 produces a corrected version of the query, which is typically successful in the first iteration. The corrected result is returned to the \texttt{Answer Generator} for further processing. Finally, the \texttt{Answer Generator} converts the query output into a coherent, natural language response, delivered to the user through the chat interface. If the query fails entirely or yields no result, the user gets informed.

\subsection{SPARQL Behavior Tree Generation Workflow}
\label{sec:sbt-factory}

\begin{figure}
\vspace{-0.7cm}
    \centering
    \includegraphics[width=0.6\linewidth]{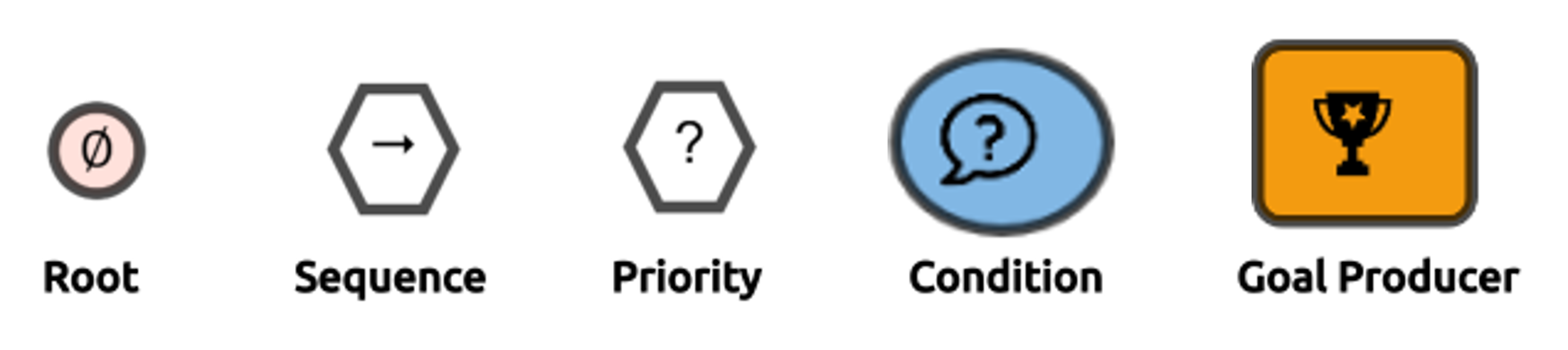}
    \vspace{-0.3cm}
    \caption{SPARQL-BT nodes that can be generated by the NLP interface}
    \label{fig:sbt-nodes}
    \vspace{-0.5cm}
\end{figure}

\noindent
The generation of SPARQL Behavior Trees in AJAN hinges on reusable building blocks—tree nodes—provided by the \emph{SBT Node Factory}. This module maintains blueprints for five primary node types: \texttt{GoalProducer}, \texttt{Condition}, \texttt{Sequence}, \texttt{Priority}, and \texttt{Root}. These nodes were specifically selected because they allow higher-level conditional behaviors to be created, which orchestrate more complex domain-integrated parameterized behaviors via the integration of additional predefined sub-behaviors via \texttt{GoalProducers}. Accordingly, a level of complexity is introduced that remains manageable even for non-experts. Each is defined via dedicated Python classes which generate the respective RDF representation needed for SBTs and designed for composability, allowing for flexible, extensible behavior models. These modular nodes enable the synthesis of executable agent logic directly from high-level user instructions. Figure~\ref{fig:sbt-nodes} illustrates them as visualized within the AJAN-Editor.

\begin{figure}[ht!]
  \centering
  \includegraphics[width=\linewidth]{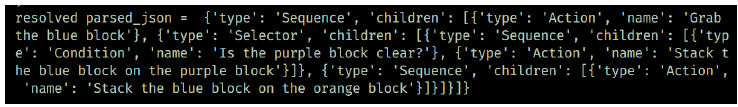}
  \caption{Generated BTF from the user's instruction}
  \label{fig:btf-example}
  \vspace{-0.5cm}
\end{figure}

\noindent
\textbf{Understanding User Instructions} SBT creation begins with parsing user input into a structured representation. This starts with generating a Behavior Tree Frame (BTF)—a scaffold devoid of attributes—which guides the SBT’s composition. Using few-shot prompting, the \texttt{gpt-3.5-turbo-16k-0613} model receives the user instruction along with 20 instruction–BTF examples and outputs a JSON-encoded frame. Figure~\ref{fig:btf-example} shows a generated BTF from the Blocks World domain, with which the blue block is placed on the purple block, otherwise on the orange one, if the purple block is not clear.
\\\\
\noindent
\textbf{Syntactic and Semantic Analysis} To enrich the BTF with logic, the system parses the instruction using \texttt{SpaCy} to extract part-of-speech tags and dependencies. Verbs, objects, and modifiers are identified for transformation into actionable semantics. Next, \texttt{gpt-3.5-turbo-16k} extracts goals and target entities from the text, forming the basis for SPARQL queries and node configuration.

\subsection{Action and Entity Linking}
\label{sec:action-entity-linking}

\begin{figure}
\centering
    \begin{subfigure}[b]{0.38\textwidth}
        \centering
        \includegraphics[width=\textwidth]{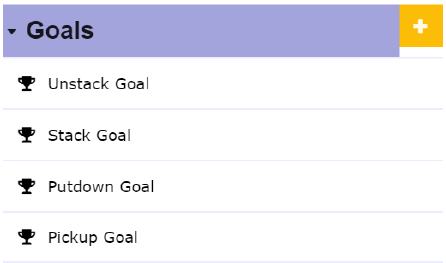}
        \caption{Blocks World AJAN goals}
        \label{fig:blocks-world-goals}
    \end{subfigure}
    \hfill
    \begin{subfigure}[b]{0.6\textwidth}
        \centering
        \includegraphics[width=\textwidth]{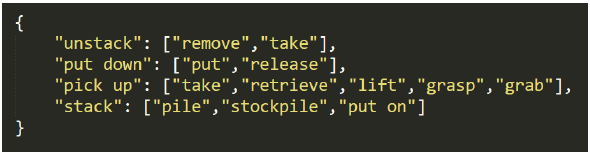}
        \caption{User dictionary module}
        \label{fig:user-dictionary}
    \end{subfigure}
\caption{AJAN goals and the respective dictionary}
\label{fig:block_user}
\vspace{-0.5cm}
\end{figure}

\noindent
After identifying actions from the user´s input, they are linked to RDF-based AJAN goals via a three-tiered mapping strategy:
\\\\
\noindent \textbf{Action Linking} First, actions are matched directly against goal labels in the RDF model. Figure~\ref{fig:blocks-world-goals} shows an example set of goals from the Blocks World agent as visualized in the AJAN-Editor. If no direct match exists, the system consults a user-specific synonym dictionary updated through previous disambiguations. Figure~\ref{fig:user-dictionary} shows a visual representation of the user dictionary module. If still unresolved, the system computes semantic similarity using the \texttt{T5} transformer model\footnote{T5: \url{https://huggingface.co/docs/transformers/model_doc/t5}}. Vectors are generated for the unknown and known actions, and cosine similarity is applied. If the similarity exceeds 0.5, the action is linked; otherwise, user disambiguation is triggered.

\noindent \textbf{Entity Linking} Entities referenced in instructions (e.g., “blue block”) must also be uniquely mapped to RDF resources. Like action linking, this process uses the \textit{Disambiguator} to resolve ambiguities. While this remains a challenging task—particularly in cases of synonymy or context-dependence—the layered strategy of direct mapping, synonym memory, and semantic similarity provides a practical solution within the AJAN ecosystem.

\subsection{Translation of User Instructions into Behavior Trees}
\label{sec:bt-construction}
\noindent
After constructing the BTF for a given user instruction, the system translates it into a complete SBT by traversing the BTF, instantiating nodes from the \textit{SBT Node Factory}, and assigning relevant properties and SPARQL queries to each node. For inner nodes that describe the internal execution logic, such as \texttt{ROOT}, \texttt{SEQUENCE}, and \texttt{PRIORITY} nodes, the factory provides predefined RDF representations that only need to be linked to child nodes in the form of an RDF list during translation. The nodes implemented that require additional parameters such as events or query definitions, are the \texttt{GoalProducer} node and the \texttt{Condition} node: \\

\begin{figure}
\vspace{-0.9cm}
\centering
    \begin{subfigure}[b]{0.56\textwidth}
        \centering
        \includegraphics[width=\textwidth]{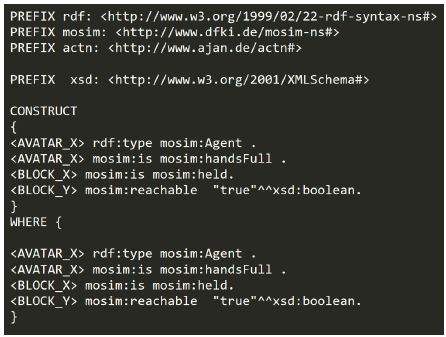}
        \caption{Construct template for the generation of GoalProducer nodes }
        \label{fig:action-template}
    \end{subfigure}
    \hfill
    \begin{subfigure}[b]{0.43\textwidth}
        \centering
        \includegraphics[width=\textwidth]{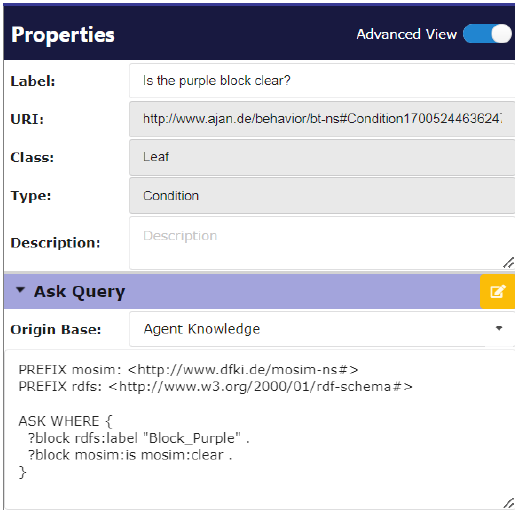}
        \caption{Condition property with generated ASK query}
        \label{fig:condition-node}
    \end{subfigure}
    \vspace{-0.5cm}
\caption{Template for GoalProducers and generated Condition properties}
\label{fig:action_blocks}
\vspace{-0.5cm}
\end{figure}

\begin{figure}
\vspace{-0.5cm}
    \centering
    \includegraphics[width=1\linewidth]{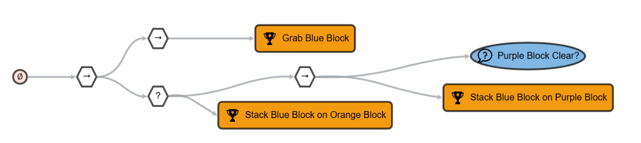}
    \vspace{-0.4cm}
    \caption{Generated SPARQL-BT for the Blocks World demo}
    \label{fig:generated-sbt}
    \vspace{-0.5cm}
\end{figure}

\noindent 
Each \textbf{GoalProducer} node corresponds to an atomic agent action or a sub-behavior tree. The target goal URI is inserted into the node based on its RDF label (see Fig.~\ref{fig:blocks-world-goals}). A SPARQL \texttt{CONSTRUCT} query template—linked to the action—is populated with previously identified entities, serving as dynamic arguments. This query encodes the precondition context required for successful execution, derived from the associated \texttt{ASK} query. Figure~\ref{fig:action-template} shows an example template for the action \texttt{stack}, with placeholders like \texttt{<BLOCK\_X>} and \texttt{<BLOCK\_Y>}. 
\\\\
\noindent 
\textbf{Condition} nodes are generated by directly translating conditional phrases into SPARQL \texttt{ASK} queries without using templates. Each query reflects the condition extracted during semantic parsing. Figure~\ref{fig:condition-node} shows a condition node as rendered in the AJAN-Editor. Details can be found in the AJAN Wiki\footnote{\url{https://github.com/aantakli/AJAN-service/wiki/SPARQL-BT-Leaf-Nodes}}.
\\\\
\noindent 
Once assembled, the SBT is either saved to the RDF repository for offline inspection in the AJAN Editor or pushed directly to the triplestore for execution in online mode. This supports both iterative development and live control. Figure~\ref{fig:generated-sbt} displays the complete SBT for the Blocks World example instruction.

\subsection{Semantic Search over AJAN Documentation}
\label{sec:semantic-search}

\noindent
To respond to documentation-related queries, the system uses a semantic search workflow that encodes textual resources (e.g., the AJAN Wiki) into dense vector representations. These vectors are indexed for similarity-based retrieval, enabling accurate extraction of relevant content during user interaction. Text is converted into fixed-size semantic vectors using OpenAI’s \texttt{text-embedding-ada-002} model\footnote{\url{https://platform.openai.com/docs/guides/embeddings}}. Each input string is encoded into a high-dimensional vector reflecting its semantic meaning, allowing similar texts to be mapped to nearby points in vector space. AJAN documentation hosted on GitHub is automatically fetched and parsed using \texttt{BeautifulSoup}\footnote{\url{https://pypi.org/project/beautifulsoup4/}}. It is segmented into paragraph- or section-level chunks, which are embedded and stored in \texttt{Pinecone}\footnote{\url{https://www.pinecone.io}}, a managed vector database. Each vector is stored with metadata—document URI, section title, and offset—to enable provenance-aware retrieval. User queries are embedded using the same model to ensure compatibility. Cosine similarity is then applied to identify the top-$K$ most relevant document vectors (typically $K=5$). The retrieved segments, along with the query, are compiled into a prompt for the \texttt{gpt-3.5-turbo-0613} model\footnote{\url{https://platform.openai.com/docs/models}}, which generates an answer grounded in the retrieved context. This response is passed to the \textit{Orchestrator} for delivery.

%% file: sections/applications.tex
\section{Application Example}

\begin{figure}[h!]
\vspace{-0.9cm}
\centering
    \begin{subfigure}[b]{0.31\textwidth}
        \centering
        \includegraphics[width=\textwidth]{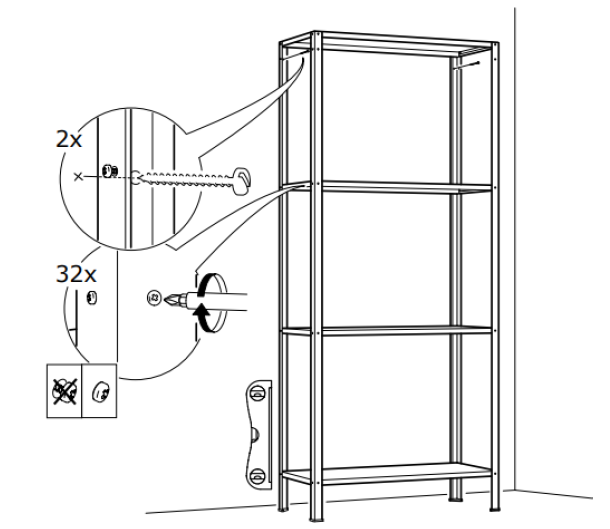}
        \caption{Shelf assembly steps}
        \label{fig:hyllis_ikea}
    \end{subfigure}
    \hfill
    \begin{subfigure}[b]{0.50\textwidth}
        \centering
        \includegraphics[width=\textwidth]{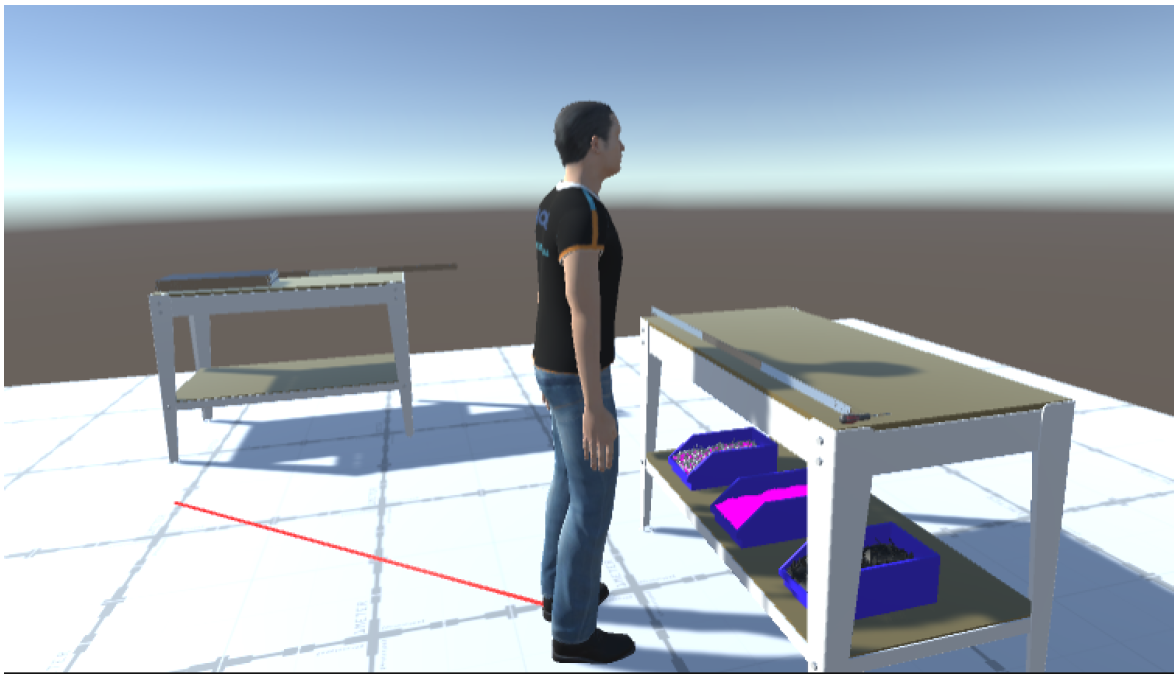}
        \caption{3D MOSIM environment}
        \label{fig:hyllis_3d}
    \end{subfigure}
    \vspace{-0.2cm}
\caption{Manual assembly application example in the Industry 4.0 context}
\label{fig:hyllis}
\vspace{-0.5cm}
\end{figure}

\noindent
As shown in \cite{antakli2023ajan}, the AJAN framework has been already applied ain various  domains. 
The goal of the AJAN-Editor is to supports users within these domains in implementing agent-based applications, particularly targeting 
developers and MAS experts in the Semantic Web. In the past, AJAN was very often used for Industry 4.0 scenarios to simulate human-in-the-loop processes in 3D environments, supporting ergonomic evaluations and feasibility studies. The following application example continues 
this line of use cases, focusing on human process planners who design and validate manual assembly steps.

\noindent 
In the application example described below, the AJAN-Editor, primarily with its LLM-based user interface, is used to model manual work steps performed by a worker and simulate them directly in 3D using the MOSIM framework \cite{mosim}. The example used is the assembly of a shelf\footnote{Shelf construction manual: \url{https://manuals.plus/ikea/hyllis-shelving-unit-in-outdoor-manual}}. In general, the assembly process is as follows: (i) Position the leg; (ii) position the shelf; (iii) insert the screw, (iv) and tighten the screw. Since the shelf has four legs and four shelves, these four steps are repeated several times. For the MOSIM scene, the AJAN agent has access to the general ontology or the T-Box and A-Box information in RDF and can therefore perform the assembly in the MOSIM environment using predefined SPARQL-BTs executing MOSIM animation calls. In \texttt{Video I}\footnote{AJAN-Editor NLP: \url{https://anonymous.4open.science/r/AJAN-material-1E45/AJAN-MOSIM.mp4}}, this process can be seen accompanied by the AJAN-Editor Live view. \texttt{Video II}\footnote{AJAN-Editor NLP: \url{https://anonymous.4open.science/r/AJAN-material-1E45/AJAN-NLP.mov}} also shows the use of the AJAN-NLP interface. The various modes of the interface are demonstrated here. In \textbf{offline mode}, a SPARQL-BT for the Blocks World demo is generated textually, stored and later on executed in a MOSIM simulation. In \textbf{offline mode}, on the other hand, the user is speaking directly to an instantiated agent to obtain information about the scene or execute commands immediately. The last part of the video demonstrates the AJAN ChatBot for the AJAN Wiki.

%% file: sections/relatedwork_short.tex
\section{Related Work}
\label{sec:related-work}

The integration of natural language processing (NLP) for Behavior Tree (BT) generation as in AJAN 
may enhance the accessibility and scalability of agent modeling.  
There are few relevant approaches in this direction. For example, Friedrich et al.~\cite{friedrich2011process} used templates 
to extract behavior components but lacked semantic and domain flexibility. 
Shu et al.~\cite{shu2019behavior} achieved domain-specific precision via rule-based methods in the medical field, without dynamic generation. 
Suddrey et al.~\cite{suddrey2022learning} introduced a modular, grammar-based generator with dialogue-driven disambiguation and a Learning Mode, yet suffered from handcrafted rule limitations and absent error handling. 
Jin and Zhou~\cite{jin2022integrating} parsed input into action sequences but did not construct full BTs. 
Cao and Lee~\cite{cao2023robot} employed LLMs to populate a fixed BT structure, showing resilience to linguistic variation but lacking editing, modularity, and disambiguation. Unlike their fine-tuning approach, we found few-shot prompting more effective for preserving reasoning abilities.

\noindent 
On the other hand, relevant work on natural language to SPARQL translation gained some attraction as well. 
Liang et al.~\cite{liang2021querying} proposed a modular, DBpedia-specific pipeline using Tree-LSTMs for query ranking. 
Sima et al.~\cite{sima2021bio} built schema-driven queries from lookup tables, limited to simple questions and scientific knowledge graphs (KG). 
Borroto et al.~\cite{borroto2021system} applied dual-LSTM models for entity extraction and template generation, constrained by outdated embeddings. 
Chen et al.~\cite{chen2019bidirectional} developed BAMnet, a bi-attentive system with strong interpretability but high resource needs. Rony et al.~\cite{rony2022sgpt} introduced SGPT, integrating KG knowledge into GPT-2, 
and Kovriguina et al. \cite{kovriguina2023sparqlgen} proposed SPARQLGEN as a further development of SGPT, both improving semantic alignment but reducing explainability and cross-KG adaptability. Our approach emphasizes modularity, KG independence, and transparent, context-aware query generation without relying on rigid or monolithic architectures.

\noindent
In general, our method overcomes prior shortcomings through modular, editable BT synthesis with integrated SPARQL generation and live agent deployment.

%% file: sections/conclusion.tex
\section{Conclusion}

\noindent
This paper introduced the AJAN-Editor with its integrated LLM-based user interface for modeling and executing Semantic Web-enabled agents, and demonstrated its use through an example 3D simulation scenario. The AJAN-Editor provides a comprehensive environment for managing agent templates, modeling, visualizing and monitoring SPARQL Behavior Trees and querying agent knowledge. The LLM interface enables users to generate executable agent behaviors and SPARQL queries from natural language, supporting both textual and vocal modalities in online and offline modes. Unlike earlier rule-based, template-driven, or limited LLM-based approaches, our system enables more modular and editable Behavior Tree construction. It also improves SPARQL generation by overcoming key limitations of previous LLM based approaches, especially by enabling knowledge graph independence. Furthermore, this is the first work to use LLMs to generate SPARQL-extendet Behavior Trees described in RDF, which can also be processed directly by an agent in its dynamic environment while lowering the barrier for developing semantic web agents, even for non-experts.